\documentclass[conference]{IEEEtran}
\IEEEoverridecommandlockouts
\usepackage{cite}
\usepackage{amsmath,amssymb,amsfonts}
\usepackage{algorithmic}
\usepackage{graphicx}
\usepackage{textcomp}
\usepackage{xcolor}
\usepackage{todonotes}
\usepackage{booktabs}
\usepackage{multirow}
\usepackage[top=0.75in, bottom=1.0in, left=0.625in, right=0.625in]{geometry}
\usepackage{pifont}

\newcommand{\cmark}{\ding{51}}
\newcommand{\xmark}{\ding{55}}
\usepackage{booktabs}

\usepackage[most]{tcolorbox}

\newtcolorbox{qbox}{
enhanced,
boxrule=0pt,frame hidden,
borderline west={4pt}{0pt}{gray!75!black},
colback=gray!10!white,
sharp corners
}

\def\BibTeX{{\rm B\kern-.05em{\sc i\kern-.025em b}\kern-.08em
    T\kern-.1667em\lower.7ex\hbox{E}\kern-.125emX}}

\begin{document}

\title{EnergyLens: Interpretable Closed-Form Energy Models for Multimodal LLM Inference Serving}

\author{\IEEEauthorblockN{1\textsuperscript{st} Vittorio Palladino}
\IEEEauthorblockA{\textit{Politecnico di Milano \& UIC}\\
vpall3@uic.edu}
\and
\IEEEauthorblockN{2\textsuperscript{nd} Gianluca Palermo}
\IEEEauthorblockA{\textit{Politecnico di Milano}\\
gianluca.palermo@polimi.it}
\and
\IEEEauthorblockN{3\textsuperscript{rd} Michael E. Papka}
\IEEEauthorblockA{\textit{University of Illinois Chicago}\\
papka@uic.edu}
\and
\IEEEauthorblockN{4\textsuperscript{th} Zhiling Lan}
\IEEEauthorblockA{\textit{University of Illinois Chicago}\\
zlan@uic.edu}
}

\maketitle
\begin{abstract}
As large language models span dense, mixture-of-experts, and state-space architectures and are deployed on heterogeneous accelerators under increasingly diverse multimodal workloads, optimising inference energy has become as critical as optimizing latency and throughput. Existing approaches either treat latency as an energy proxy or rely on data-hungry black-box surrogates. Both fail under varying parallelism strategies: latency and energy optima diverge in over 20\% of configurations we tested, and black-box surrogates require hundreds of profiling samples to generalize across model families and hardware. 
We present \textsc{EnergyLens}, which uses symbolic regression as a structure-discovery tool over profiling data to derive a single twelve-parameter closed-form energy model expressed in terms of system properties such as degree of parallelism, batch size, and sequence length.
Unlike black-box surrogates, \textsc{EnergyLens} decouples tensor and pipeline parallelism contributions and separates prefill from decode energy, making its predictions physically interpretable and actionable.
 Fitted from as few as 50 profiling
measurements, \textsc{EnergyLens} achieves 88.2\% Top-1 configuration selection accuracy
across many evaluation scenarios compared to 60.9\% for the closest prior analytical
baseline, matches the predictive accuracy of ensemble ML methods with $10\times$ fewer
profiling samples, and extrapolates reliably to unseen batch sizes and hardware platforms
without structural modification, making it a practical, interpretable tool for
energy-optimal LLM deployment.
\end{abstract}

\section{Introduction}
\label{sec:introduction}

Large language models now span dense transformers, mixture-of-experts (MoE),
and state-space architectures, are deployed across a heterogeneous landscape
of accelerators (NVIDIA, Intel, AMD), and serve increasingly diverse
multimodal workloads spanning text, image, and video. 
Inference energy has emerged as a first-class optimization objective alongside
latency and throughput~\cite{brown2020language}. Data centers running LLM inference at
scale face mounting pressure to reduce their energy footprint, yet lack reliable
tools to predict how deployment choices parallelism strategy, batch size, sequence
length affect energy consumption before committing to a
configuration.

Predicting inference energy is challenging for three primary reasons.
First, energy consumption arises from a complex interplay among model architecture, workload characteristics, and parallel execution strategy. Tensor parallelism (TP) and pipeline parallelism (PP) affect GPU utilization and idle power in fundamentally different ways that a single model cannot capture. Moreover, energy does not scale linearly with these factors, requiring specialized models for each combination of architecture, workload, and hardware.
Second, existing machine learning surrogate approaches~\cite{cheng2025energy} achieve reasonable accuracy but rely on black-box models (e.g., neural networks or tree ensembles). These methods require extensive profiling data, are tightly coupled to specific model–hardware configurations, are costly to retrain as systems evolve, and offer limited interpretability.
Third, analytical approaches that estimate energy via latency~\cite{maveriq2025} introduce systematic error when latency and energy objectives diverge a common scenario under varying parallelism configurations and multimodal workloads. In such cases, energy penalties of up to 79.8\% have been observed relative to the true energy-optimal configuration~\cite{wang2025systematic,maliakel2025tradeoffs,ozcan2025simulations}.

We address these limitations with \textsc{EnergyLens}, an energy modeling framework that leverages symbolic regression to discover compact, closed-form expressions capturing the nonlinear, physically grounded relationships governing GPU energy consumption during inference. Rather than fitting an opaque predictor to profiling data, symbolic regression searches the space of mathematical expressions and returns interpretable formulas whose terms correspond directly to measurable system properties such as tensor-parallelism degree, pipeline-bubble overhead, and per-device idle power making predictions both auditable and actionable.

Our central observation is that inference energy exhibits consistent mathematical structure across model families, modalities, hardware platforms, and that symbolic regression is well suited to uncover it. To obtain the structured measurements required to derive these formulas, we develop a profiling framework that systematically sweeps parallelism strategies, quantization schemes, batch sizes, and sequence lengths across NVIDIA, Intel, and AMD GPUs.

\begin{figure*}[t]
    \centering
    \includegraphics[width=\textwidth]{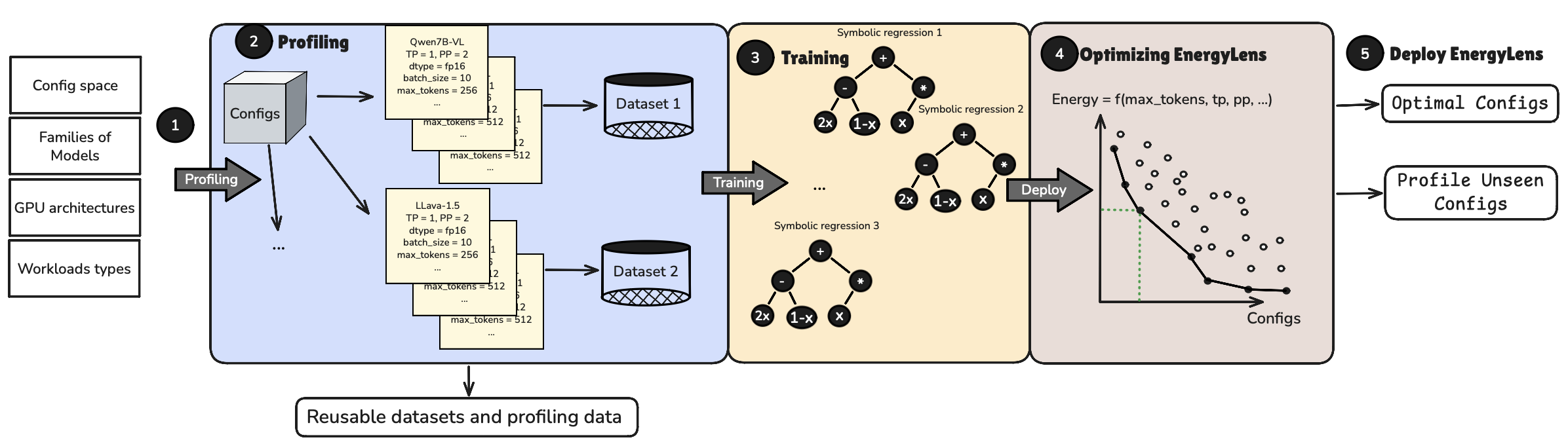}
    \caption{Overview of the \textsc{EnergyLens} training pipeline. The procedure consists
    of five stages: (1)~definition of the input space, comprising configuration parameters,
    families of models, GPU architectures, and workload types; (2)~systematic energy
    profiling of multimodal LLM serving across diverse models and configurations,
    producing reusable datasets; (3)~symbolic-regression-driven model construction
    that discovers closed-form energy formulas from profiling data; (4)~optimization
    of \textsc{EnergyLens} by identifying Pareto-optimal configurations from the learned
    energy function \( \text{Energy} = f(\texttt{max\_tokens}, \texttt{tp}, \texttt{pp}, \ldots) \) after having it being fitted on a small set of configs profiled;
    and (5)~deployment, yielding optimal configurations for known setups and
    energy profiles for unseen configurations.}
    \label{fig:framework_training}
\end{figure*}

We evaluate \textsc{EnergyLens} on a suite of text, vision-language, and video
workloads spanning dense transformers, mixture-of-experts models, and state-space models
across three GPU vendors. \textsc{EnergyLens} achieves 88.2\% Top-1 configuration
selection accuracy across many scenarios compared to 60.9\% for the closest prior
analytical baseline, matches the predictive accuracy of ensemble ML methods while
requiring $10\times$ fewer profiling samples (13.0\% MAPE at $n{=}50$ vs.\ 49.0\% for
Random Forest). The formula transfers across model families and hardware platforms without
structural modification: only its scalar coefficients need to be re-fitted on a small
representative set of measurements. 
Our contributions are as follows:

\begin{enumerate}
\item  We develop \textsc{EnergyLens}, \emph{a unified closed-form energy model} that decouples the contributions of tensor and pipeline parallelism, batch size, and sequence length, with parameters that carry explicit physical meaning and are fitted accurately from as few as 50 profiling measurements.
\item  We demonstrate that \emph{symbolic
regression} can discover nonlinear formula structure that generalizes across
diverse architectures (dense, MoE, SSM), modalities (text, image, video), and
GPU platforms (NVIDIA, Intel, AMD) without changing the formula template.
\item We implement \emph{a profiling framework} that systematically explores the joint space of parallelism
strategies, quantization schemes, batch sizes, and sequence lengths, producing
the structured energy measurements needed to fit and validate analytical models.
\item By modeling energy directly rather than using latency as a proxy, ENERGYLENS identifies more energy-efficient parallelism configurations consistently outperforming prior analytical baselines, and ML surrogates while requiring substantially fewer profiling samples.
\end{enumerate}

The remainder of this paper is organized as follows. \S\ref{sec:related-work} reviews
related work on analytical, surrogate, and latency-based energy models.
\S\ref{sec:methods} formalises the problem and characterises the deployment space.
\S\ref{subsec:formulation} describes the \textsc{EnergyLens} framework, covering the profiling
harness, symbolic regression pipeline, and unified energy formula.
\S\ref{sec:experiments} presents the experimental evaluation, including serving-parameter
characterisation and predictive benchmarks against all baselines.
\S\ref{sec:conclusion} concludes and outlines directions for future work.


\section{Related Work}
\label{sec:related-work}

We organize prior work along four axes: analytical and roofline-based
models, black-box ML surrogates, energy-aware deployment systems, and
environmental-footprint estimation models.
Table~\ref{tab:related} summarizes how \textsc{EnergyLens} compares to the
closest prior studies.

\begin{table*}[t]
\centering
\caption{Comparison of \textsc{EnergyLens} with representative prior
energy and performance models for LLM inference.
\cmark{}/\xmark{} indicate whether the property is supported.}
\label{tab:related}
\small
\begin{tabular}{lccccc}
\toprule
Method & Interpretable & Decouples TP/PP & Multi-GPU & Multimodal & Cross-vendor \\
\midrule
Roofline~\cite{williams2009roofline}             & \cmark & \xmark & \xmark & \xmark & \cmark \\
\textsc{SweetSpot}~\cite{cavagna2025sweetspot}   & \cmark & \xmark & \xmark & \xmark & \xmark \\
\textsc{MaverIQ}~\cite{maveriq2025}              & \cmark & \xmark & \cmark & \xmark & \xmark \\
PIE-P~\cite{dutt2025piep}                        & \cmark & \cmark & \cmark & \xmark & \xmark \\
ML surrogates~\cite{cheng2025energy}             & \xmark & \cmark & \cmark & \xmark & \xmark \\
LLMCO2~\cite{fu2025llmco2}                       & \xmark & \xmark & \xmark & \cmark & \cmark \\
Prompt-level estimators~\cite{prompts2power2025,hungryai2025} & \xmark & \xmark & \xmark & \cmark & \cmark \\
\textbf{\textsc{EnergyLens} (ours)}              & \cmark & \cmark & \cmark & \cmark & \cmark \\
\bottomrule
\end{tabular}
\end{table*}

\subsection{Analytical and Roofline-Based Models}

Classical roofline models~\cite{williams2009roofline} bound compute
throughput via arithmetic intensity and peak memory bandwidth. They treat
parallelism as a scalar multiplier and cannot differentiate between TP and
PP, whose energy effects are qualitatively distinct
(\S\ref{subsec:profiling-results}).

\textsc{MaverIQ}~\cite{maveriq2025} derives an analytical latency
formula from first principles and is the closest prior baseline to our
work. However, it first targets latency rather than energy, implicitly
assuming the two share an optimum, secondly evaluates only on text-only workloads on a single
NVIDIA platform, and lastly models batch-size scaling as linear, which
breaks at the saturation regime our profiling consistently
exposes (\S\ref{subsubsec:batch_size_energy}).

\textsc{SweetSpot}~\cite{cavagna2025sweetspot} derives a closed-form
efficiency curve from Transformer FLOPs and memory-access complexity,
correctly identifying non-linear sweet spots as a function of sequence
length on a single GPU. It does not address multi-GPU parallelism or
multimodal workloads, and is therefore not directly comparable on the
deployment regimes we target.

PIE-P~\cite{dutt2025piep} decomposes multi-GPU inference energy into
compute, communication, and synchronisation components under TP and PP.
Conceptually it is close analytically to ours work, but it requires
per-operator instrumentation, does not cover multimodal inputs, and has
been validated only on a single GPU vendor.

\textsc{EnergyLens} (ours) targets the same multi-GPU parallelism regime as PIE-P while matching the interpretability of \textsc{SweetSpot}, requires no per-operator instrumentation, fits from $\sim$50 profiling samples, and generalizes across GPU vendors and input modalities without structural modification.

\subsection{Black-Box Surrogate Models}

Data-driven surrogates random forests, gradient-boosted trees, and neural
networks~\cite{cheng2025energy} achieve strong in-distribution accuracy
but share three weaknesses for our setting.

First, they are \emph{data-hungry}: a Random Forest needs $n{=}500$
measurements to reach 12.3\% MAPE on Mistral-7B, whereas
\textsc{EnergyLens} achieves 13.0\% MAPE at $n{=}50$, a $10\times$
reduction in profiling cost as shown in \ref{subsubsec:sample-efficiency}

Second, they are \emph{black box models}: hundreds of decision trees provide no
physical insight into why one configuration is more efficient than another.

Third, they \emph{extrapolate poorly}: when trained on the smallest
40\% of batch sizes and evaluated on larger ones, ensemble methods
degrade sharply, while \textsc{EnergyLens} maintains $\geq93\%$
pairwise ranking accuracy on five of seven datasets, as shown later.

A practical limitation is that black-box surrogates require retraining from scratch for each new model–hardware combination. In contrast, because the structural form of \textsc{EnergyLens}'s formula is fixed, only its scalar coefficients need to be re-fitted.

\subsection{Energy-Aware Deployment and Environmental Footprint}
Complementary work improves inference energy by modifying \emph{how} requests
are executed: speculative decoding~\cite{leviathan2023speculative}, KV-cache
compression, and dynamic batching reduce per-request cost without modelling
energy directly, while \textsc{DynamoLLM}~\cite{dynamollm2025}
dynamically adjusts instance count, GPU frequency, and model parallelism under
changing load.
A broader sustainability literature estimates prompt-level energy, carbon, and
resource footprint for AI inference, including \emph{From Prompts to
Power}~\cite{prompts2power2025}, \textsc{LLMCO2}~\cite{fu2025llmco2},
\textsc{Clover}~\cite{clover2023}, and infrastructure-scale studies such as
\emph{How Hungry is AI?}~\cite{hungryai2025}.
Both lines of work are orthogonal to \textsc{EnergyLens}: they either modify
runtime execution or estimate pointwise footprints, whereas our model exposes a
parametric closed-form expression over the full deployment parameter space,
re-fittable for unseen model--hardware pairs.

\section{Methodology}
\label{sec:methods}

This section formalizes the energy prediction problem and describes the orthogonal axes of the deployment space.

\subsection{Problem Formulation}
\label{sec:problem-formulation}

We consider a serving request characterised by a tuple
$(\mathcal{M}, \mathcal{W}, \mathcal{H}, \boldsymbol{\theta})$, where
$\mathcal{M}$ denotes the model architecture, $\mathcal{W}$ the workload
modality, $\mathcal{H}$ the hardware platform, and $\boldsymbol{\theta}$
the vector of serving-engine parameters (parallelism degrees, batch size,
sequence length, quantisation scheme). Our goal is to learn a compact,
closed-form function
\begin{equation}
\widehat{E} = f\bigl(\mathcal{M}, \mathcal{W}, \mathcal{H},
\boldsymbol{\theta};\, \boldsymbol{\phi}\bigr),
\label{eq:problem}
\end{equation}
parametrised by a small fitted vector $\boldsymbol{\phi}$, that predicts
the per-inference energy consumption $E$ (in Joules) while satisfying four
properties:
\begin{enumerate}
    \item[\textbf{P1}] \emph{Parallelism decoupling.} The function must
    distinguish the contributions of tensor parallelism (TP) and pipeline
    parallelism (PP), which have qualitatively different physical effects
    on per-GPU computation and idle power.
    \item[\textbf{P2}] \emph{Phase decomposition.} The function must
    separate the compute-bound prefill phase from the memory-bound decode
    phase, since their energy scales with different combinations of
    workload parameters.
    \item[\textbf{P3}] \emph{Cross-domain transferability.} The functional
    form must generalise across model families, modalities, and GPU
    vendors with only the scalar parameters $\boldsymbol{\phi}$ re-fitted.
    \item[\textbf{P4}] \emph{Sample efficiency.} The parameters
    $\boldsymbol{\phi}$ must be estimable from a small profiling budget compatible with production constraints.
\end{enumerate}
We define an \emph{energy-optimal configuration} as
$\boldsymbol{\theta}^{\star} = \arg\min_{\boldsymbol{\theta} \in \Theta}
E(\mathcal{M}, \mathcal{W}, \mathcal{H}, \boldsymbol{\theta})$, where
$\Theta$ is the discrete configuration space defined in
\S\ref{sec:problem-space}. The downstream objective of \textsc{EnergyLens}
is to identify $\boldsymbol{\theta}^{\star}$ using $\widehat{E}$ as a
selection oracle, so we evaluate not only predictive error
(\S\ref{subsubsec:sample-efficiency}) but also \emph{ranking} and \emph{selection}
quality (\S\ref{subsubsec:pairwise-ranking}).

\subsection{Deployment Space}
\label{sec:problem-space}

The deployment space is organized along four orthogonal axes, summarized in
Table~\ref{tab:deployment-space}, each of which influences energy in
qualitatively different ways.

\begin{table}[t]
\centering
\caption{The four orthogonal axes of the deployment space.}
\label{tab:deployment-space}
\scriptsize
\setlength{\tabcolsep}{4pt}      
\renewcommand{\arraystretch}{0.9} 

\begin{tabular}{lp{7.2cm}}
\toprule
Axis & Values \\
\midrule

Model architecture &
Dense (Mistral-7B, Qwen2-VL-7B, Nemotron-v2-9B), MoE (Qwen1.5-MoE-A2.7B), SSM (Mamba-130M) \\

Workload modality &
Text (WikiText-2), Image+text (COCO-2017), Video+text (Video-MME), Video-only (MSR-VTT) \\

Hardware platform &
NVIDIA A100, AMD MI100, Intel GPU Max 1550 \\

Serving parameters &
TP, PP, batch size, max output tokens (Table~\ref{tab:vllm-params}) \\

\bottomrule
\end{tabular}
\end{table}

\subsubsection{Model Architecture}
We consider three families that span the dominant design philosophies of
current LLM deployment:
\begin{itemize}
    \item \emph{Dense Transformer} (Mistral-7B-Instruct-v0.3,
    Qwen2-VL-7B-Instruct, NVIDIA Nemotron-v2-9B): every token attends to
    every other token via multi-head self-attention, and all feed-forward
    parameters activate per token. Per-token compute is constant, but
    memory bandwidth scales with sequence length due to KV-cache growth.
    \item \emph{Mixture-of-Experts} (Qwen1.5-MoE-A2.7B): the feed-forward
    block is a set of expert sub-networks, only a small subset of which is
    activated per token via a learned router. This decouples total
    parameter count from per-token FLOPs but introduces all-to-all
    communication overhead under parallelism.
    \item \emph{State-Space Model} (Mamba-130M): a recurrent architecture
    based on selective state-space models that replaces attention with a
    structured recurrence over a fixed-size hidden state, reducing
    per-step complexity from $\mathcal{O}(n^2)$ to $\mathcal{O}(n)$ and
    eliminating the KV-cache.
\end{itemize}

\subsubsection{Workload Modality}
We consider four modalities with progressively increasing computational and memory demands:
\begin{itemize}
    \item \emph{Text-only} (WikiText-2-raw-v1): pure language modelling.
    \item \emph{Image+text} (COCO 2017): vision-language inference
    requiring a vision-encoder forward pass and additional visual tokens
    during prefill.
    \item \emph{Video+text} (Video-MME): temporal visual understanding
    requiring encoding of multiple frames and substantially longer
    effective sequence lengths.
    \item \emph{Video-only} (MSR-VTT): video understanding without
    accompanying text context.
\end{itemize}

\subsubsection{Hardware Platform}
To assess generalization across vendors, we cover hardware accelerators
from all three major GPU vendors:
\begin{itemize}
    \item \emph{NVIDIA A100}: part of the Sophia system at
          ALCF~\cite{alcf_sophia}, with each compute node equipped with
          $4{\times}$ 40\,GB A100 GPUs interconnected via NVLink~3.0.

    \item \emph{Intel Data Center GPU Max 1550}: part of the Aurora system
          at ALCF~\cite{alcf_aurora_arxiv}, with each compute node equipped
          with $6{\times}$ 128\,GB Ponte Vecchio accelerators using Intel
          Xe Link fabric.

    \item \emph{AMD Instinct MI100}: part of ChameleonCloud, with compute
          nodes equipped with $2{\times}$ 192\,GB MI100 GPUs connected via
          AMD Infinity Fabric~\cite{chameleon_acm23}.
\end{itemize}

\subsubsection{Serving-Engine Parameters}
\label{sec:serving-params}
In this study, we use the vLLM serving engine. We partition its configuration parameters into two classes based on the cost of applying a change (Table~\ref{tab:vllm-params}):
\begin{itemize}
    \item \emph{Heavy configurations} require a full model reload and
    determine the deployment architecture: TP degree
    $d_{\mathrm{TP}} \in \{1, 2, 4\}$, PP degree
    $d_{\mathrm{PP}} \in \{1, 2, 4\}$, and maximum number of batched
    tokens $\in \{1024, 2048, 4096, 8192, 16384\}$.
    \item \emph{Light configurations} can be modified at runtime without
    restarting the serving instance and primarily affect sampling
    behaviour: maximum output tokens $\in \{64, 128, 256, 512\}$.
\end{itemize}
This distinction has practical implications for profiling: heavy-configuration
changes must be executed as separate jobs, whereas light-configuration
sweeps can run within a single instance, substantially reducing total
profiling time.

\begin{table}[t]
\centering
\caption{vLLM configuration parameters and swept values.}
\label{tab:vllm-params}
\small
\begin{tabular}{lll}
\toprule
Class & Parameter & Values \\
\midrule
\multirow{3}{*}{Heavy}
 & Tensor Parallel Size ($d_{\mathrm{TP}}$) & 1, 2, 4 \\
 & Pipeline Parallel Size ($d_{\mathrm{PP}}$) & 1, 2, 4 \\
 & Max Batched Tokens & 1024, 2048, \\
  &                   &4096, 8192, 16384 \\
\midrule
Light & Max Output Tokens & 64, 128, 256, 512 \\
\bottomrule
\end{tabular}
\end{table}

Throughout this paper, we adopt the following conventions to avoid overloading the term \emph{architecture}: \emph{platform} refers to the hardware vendor (NVIDIA, Intel, AMD); \emph{model family} refers to the neural-network type (dense Transformer, MoE, SSM); and \emph{workload} refers to the request modality (text, image+text, video+text).

\section{The \textsc{EnergyLens} Model}
\label{subsec:formulation}

\subsection{Unified Parallelism-Aware Energy Model}

We conduct extensive experiments across the different axes of the deployment space listed in Table~\ref{tab:deployment-space}. A consistent pattern emerges across independent runs spanning different model families and workload modalities. For text-only workloads, the discovered formulas consistently express energy as a function of the ratio between token load and a parallelism-weighted batch term:

\begin{equation}
  \label{eq:sr-text}
  E \;\propto\;
  \frac{c_1}{\,\text{parallelism} \cdot \bigl(\text{batch\_size}\,/\,\text{max\_tokens}\bigr) + c_2\,}
  + c_3,
\end{equation}
while for video and image workloads the search converges to a logarithmic
compression of a similar ratio:
\begin{equation}
  \label{eq:sr-video}
  E \;\propto\;
  \log\!\left(
    \frac{\text{max\_tokens}}
         {d_{\text{TP}} \cdot \text{total\_input\_tokens} \cdot c_1}
    \cdot \frac{1}{\text{parallelism}}
    + c_2
  \right).
\end{equation}
Two structural invariants recur across both modalities and all model families:
(i)~\textbf{token load over parallelism} energy scales with the number of
tokens each device must process, captured as token counts divided by a
parallelism factor; and
(ii)~\textbf{batch-size saturation} increasing batch size reduces
per-request energy with diminishing returns, encoded as batch\_size appearing
in the denominator with a sublinear or ratio-based role.

A key finding from our profiling study is that tensor parallelism~(TP) and
pipeline parallelism~(PP) affect fundamentally differently the energy and
must therefore be modeled separately.
TP splits tensor operations within each layer across GPUs, reducing per-GPU
computation through efficient tensor-core utilisation while introducing
all-reduce synchronisation overhead~\cite{megatron2021}.
PP distributes layers sequentially across GPUs, incurring pipeline-bubble
overhead and keeping all GPUs powered during idle
stages~\cite{gpipe2019,pipedream2019}.
Treating these two axes uniformly, as prior work does, combine their distinct
physical mechanisms and leads to systematic mispredictions.

We employ \emph{symbolic regression} to derive a unified, closed-form energy model. Importantly, its role is to uncover recurring structural patterns across independent runs. The resulting motifs token-load scaling, batch-size saturation, and TP/PP asymmetry are then consolidated into a single closed-form formula that preserves their structure while remaining compact and physically grounded. Furthermore, LLM inference operates in two distinct phases with different computational profiles: a compute-bound prefill phase (processing input tokens) and a memory-bound decode phase (generating output tokens). Equation~\eqref{eq:unified-tp-pp} encodes these shared motifs directly, decomposing energy into prefill, decode, and baseline overhead components:
\begin{align}
  \label{eq:unified-tp-pp}
  E &= \underbrace{\alpha_p \cdot \frac{\text{total\_input\_tokens}}
      {\bigl(\text{batch\_size}^{\beta_p} + \epsilon_p\bigr)}
      \cdot d_{\text{TP}}^{\,\gamma_{1,p}} \cdot d_{\text{PP}}^{\,\gamma_{2,p}}
      }_{\text{Prefill Energy ($E_{\text{pf}}$)}} \nonumber \\
    &\quad + \underbrace{\alpha_d \cdot \frac{\text{max\_tokens}}
      {\bigl(\text{batch\_size}^{\beta_d} + \epsilon_d\bigr)}
      \cdot d_{\text{TP}}^{\,\gamma_{1,d}} \cdot d_{\text{PP}}^{\,\gamma_{2,d}}
      }_{\text{Decode Energy ($E_{\text{dc}}$)}} \nonumber \\
    &\quad + \underbrace{\delta_1 \cdot d_{\text{TP}}
      + \delta_2 \cdot d_{\text{PP}}}_{\text{Baseline Overhead}},
\end{align}
where $d_{\text{TP}}$ and $d_{\text{PP}}$ denote the degrees of tensor and
pipeline parallelism, respectively.

These twelve parameters carry explicit physical meaning. The $\alpha_p$ and $\alpha_d$ parameters are global workload-scaling constants for the prefill and decode phases, absorbing model size and hardware throughput.
The $\beta_p$ and $\beta_d$ terms control the rate at which batching amortises per-request costs in their respective phases, while $\epsilon_p$ and $\epsilon_d$ regularise behaviour at small batch sizes.
The $\gamma$ parameters independently capture the energy impact of scaling TP ($\gamma_{1,p}, \gamma_{1,d}$) versus PP ($\gamma_{2,p}, \gamma_{2,d}$), cleanly decoupling how compute and memory bottlenecks shift under different parallelism strategies. Finally, $\delta_1$ and $\delta_2$ account for the hardware-level baseline power draw and static overheads introduced by each parallelism axis.

\subsection{Using \textsc{EnergyLens} in Practice}
\label{subsec:deployment-example}

To illustrate how \textsc{EnergyLens} is used in practice, consider an operator seeking to minimize inference energy for a vision-language model on a previously unseen hardware platform. The operator first collects a small profiling dataset of approximately 50 measurements by sweeping a representative subset of the configuration space defined in Table~\ref{tab:vllm-params}, a process that typically completes within minutes without disrupting production workloads. The twelve parameters of Eq.~\eqref{eq:unified-tp-pp} are then fitted to these measurements via L-BFGS-B optimization~\cite{lbfgsb1995}, yielding a closed-form energy model ready for deployment. The resulting formula can then be evaluated analytically over the full configuration space including parallelism degrees, batch sizes, and sequence lengths beyond those observed during profiling to identify the energy-optimal deployment configuration without exhaustive measurement. Because the formula structure is fixed across model families and hardware platforms, the same workflow applies when migrating to new models or GPUs, requiring only re-fitting of scalar coefficients on a small representative sample, additional details on sampling techniques and ablations are provided in \S\ref{subsubsec:cross-hardware}.

\section{Experiments}
\label{sec:experiments}

We evaluate \textsc{EnergyLens} along two complementary dimensions: (1) analyzing how serving parameters affect energy and latency across models, workloads, and hardware, and (2) benchmarking \textsc{EnergyLens} against baselines in terms of accuracy, sample efficiency, and configuration-selection quality.

\subsection{Experimental Setup}
\label{subsec:exp-setup}

\paragraph{Baseline methods.}
We compare \textsc{EnergyLens} against four baselines:
\begin{itemize}
    \item \textbf{Linear Regression}~\cite{galton1886regression}: a standard linear model over the same input features, representing the simplest parametric baseline.
    \item \textbf{Random Forest}~\cite{breiman2001random}: an ensemble of 100 decision trees, representing a nonparametric, data-hungry approach.
    \item \textbf{Gradient Boosting}~\cite{friedman2001greedy}: a sequential ensemble method that iteratively corrects residual errors, representing the strongest ML baseline.
    \item \textbf{\textsc{MaverIQ}}~\cite{maveriq2025}: the closest prior analytical approach, which derives a formula for end-to-end request latency from profiling data. Since \textsc{MaverIQ} targets latency rather than energy; we multiply its predicted latency by the GPU power at the corresponding configuration and re-fit its parameters via gradient descent to minimize energy MAPE, following the methodology described in their paper and code.
\end{itemize}

\paragraph{Evaluation metrics.}
We report three metrics:
\begin{itemize}
    \item \textbf{MAPE} (Mean Absolute Percentage Error): scale-invariant accuracy, our primary metric.
    \item \textbf{$R^2$} (Coefficient of Determination): fraction of variance explained.
    \item \textbf{RMSE} (Root Mean Squared Error): absolute prediction error in Joules.
\end{itemize}

\paragraph{Datasets and models}
We evaluate five hardware–model combinations: Mamba-130M, Nemotron-v2-9B, and Mistral-7B on NVIDIA A100; Mistral-7B on Intel GPU Max 1550; and Mistral-7B on AMD MI100. For multimodal evaluation, we additionally profile Qwen2-VL-7B-Instruct under text-only, image+text, and video+text workloads.

\subsection{Empirical Foundations of \textsc{EnergyLens}
}
\label{subsec:profiling-results}

Before evaluating \textsc{EnergyLens} as a predictive model, we present the profiling results that informed its design. These experiments reveal how serving parameters parallelism strategy, batch size, and input modality affect both energy and latency, and critically demonstrate that latency cannot always be used as a reliable proxy for energy optimization.

\subsubsection{Latency vs.\ TP/PP Configuration}
\label{subsubsec:tp_pp_latency}

\begin{figure}[htbp]
    \centering
    \includegraphics[width=\columnwidth]{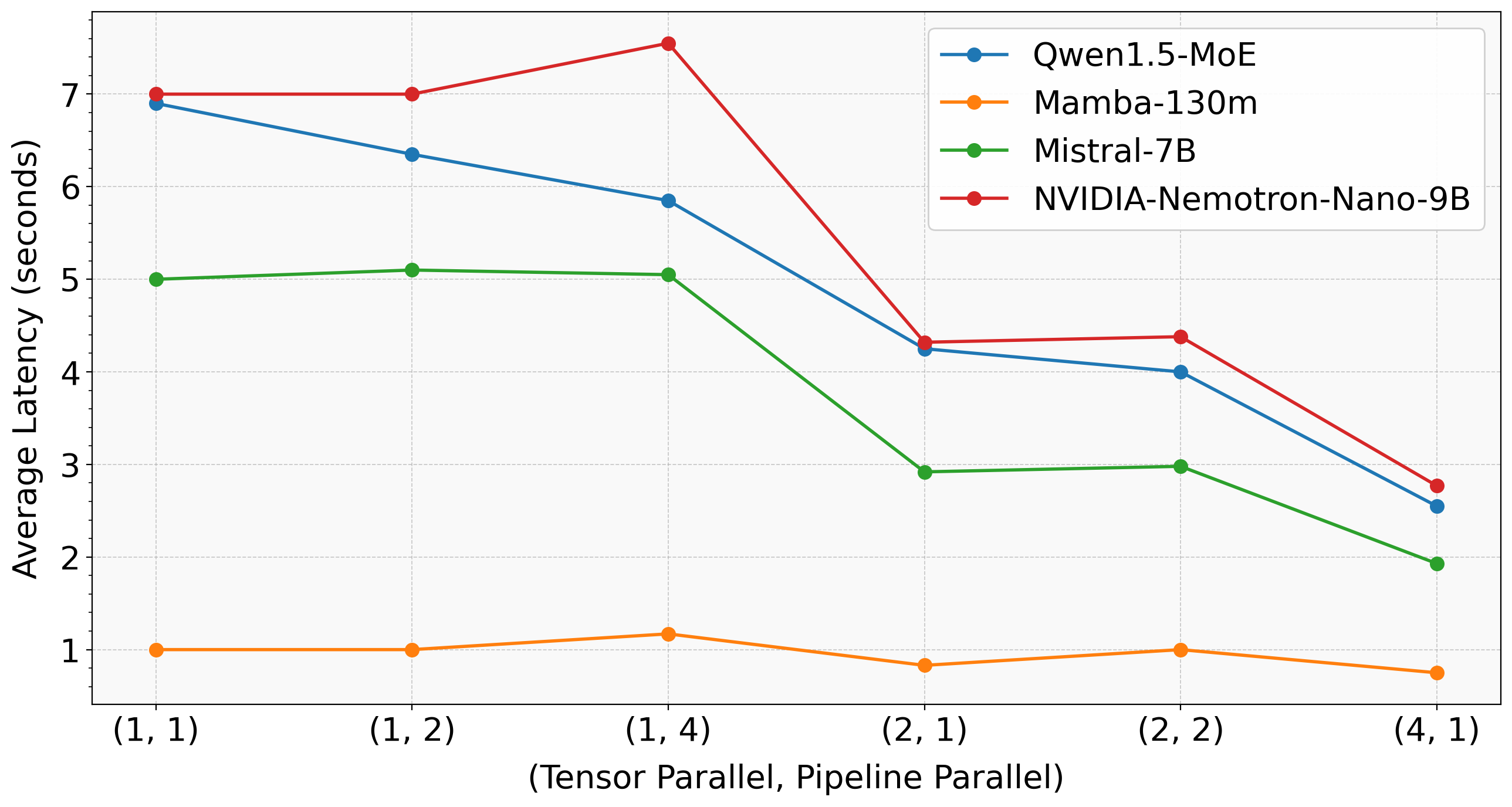}
    \caption{Average latency (seconds) versus TP/PP configuration for four
    model architectures: Qwen1.5-MoE (blue), Mamba-130m (orange), Mistral-7B
    (green), and NVIDIA Nemotron-Nano-9B (red). Dense and MoE models achieve
    60--64\% latency reductions from $\text{TP}=1$ to $\text{TP}=4$, while
    Mamba-130m remains largely insensitive due to its cache-free recurrent
    architecture.}
    \label{fig:tp_pp_latency_avg}
\end{figure}

Figure~\ref{fig:tp_pp_latency_avg} shows average latency as a function of
TP/PP configuration for four models spanning three architecture families on
NVIDIA A100. The curves separate into three distinct regimes that remain
stable across all parallelism settings, with TP driving the dominant latency
reductions for all transformer-based models and PP having negligible or even
detrimental effect.

Dense and MoE transformers respond strongly to tensor parallelism because it
partitions both attention weight matrices and the KV-cache, easing the
memory-bandwidth bottleneck that governs decode. Pure pipeline parallelism is
at best neutral for these models: Nemotron's latency actually increases under
PP=4, as inter-stage communication and bubble overhead exceed any benefit from
distributing parameters.

Qwen1.5-MoE alone derives non-trivial benefit from PP, which we attribute to
the alignment between expert sharding and pipeline stages: distributing experts
reduces per-stage memory footprint without the heavy all-reduce traffic seen in
dense TP. Mamba-130m is insensitive to both axes the absence of a KV-cache
removes the memory pressure TP typically relieves, and the model's small size
fits within a single A100, removing any motivation for PP.

A practical consequence is that the latency gap between architectures collapses
substantially under aggressive TP. Parallelism strategy can therefore
reshape rather than merely scale the relative cost ordering of candidate
models.

\subsubsection{Energy vs.\ TP/PP Configuration}
\label{subsubsec:tp_pp_energy}

\begin{figure}[htbp]
    \centering
    \includegraphics[width=\columnwidth]{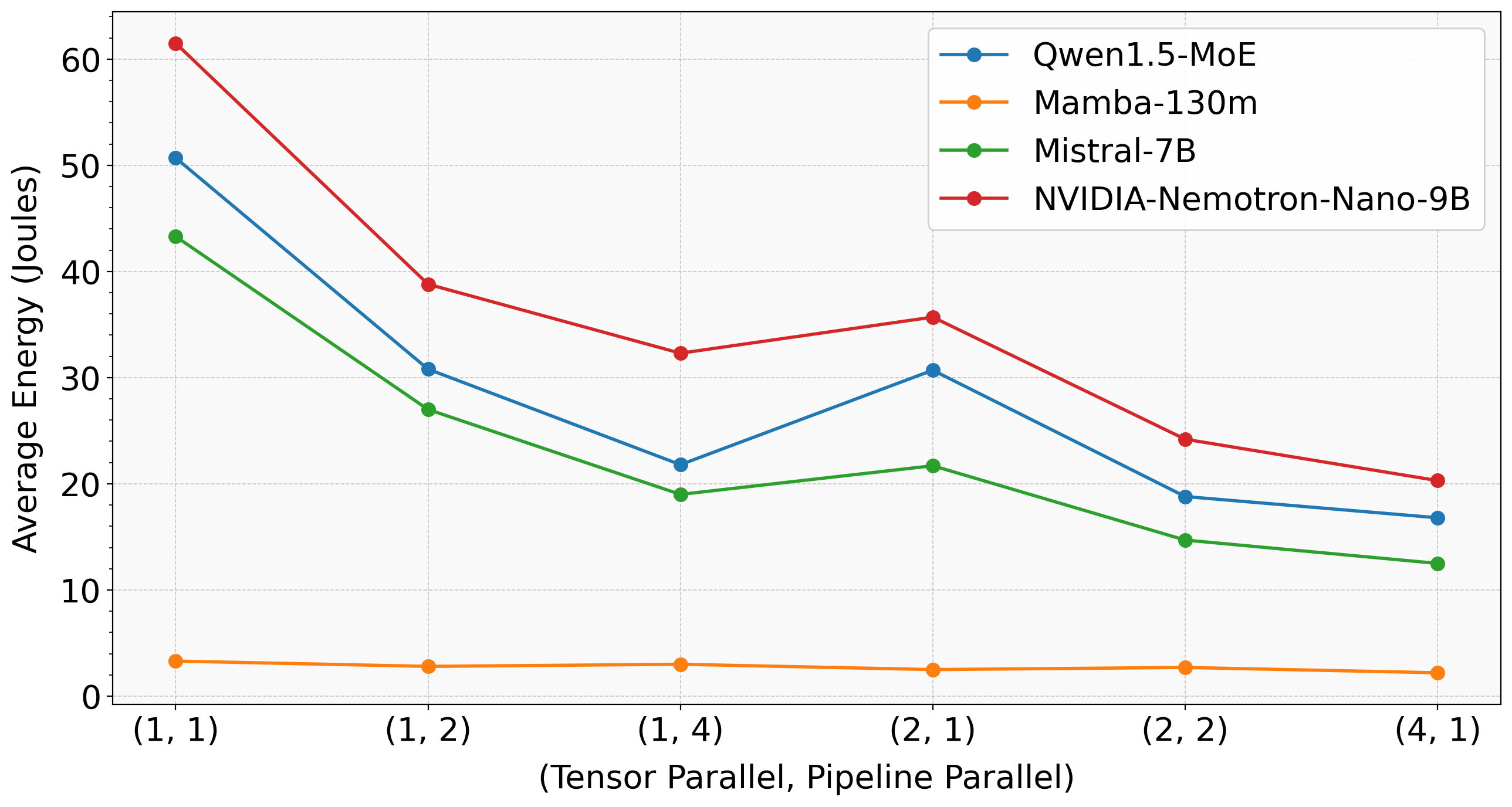}
    \caption{Average energy (Joules) versus TP/PP configuration for four model
    architectures. Energy trends largely follow latency, with all large models
    achieving 67--71\% reductions at $(\text{TP}{=}4, \text{PP}{=}1)$.
    Crucially, unlike latency, pipeline parallelism provides measurable energy
    savings even for dense transformers. Mamba-130m maintains 10--20$\times$
    lower energy across all configurations.}
    \label{fig:tp_pp_energy_avg}
\end{figure}

Figure~\ref{fig:tp_pp_energy_avg} shows energy consumption across the same
configurations. The overall shape mirrors the latency plot, but two divergences
reveal why latency alone is an insufficient proxy for energy optimization.

First, pipeline parallelism yields meaningful energy savings for dense
transformers both Nemotron-Nano-9B and Mistral-7B show substantial drops
under PP-only configurations despite providing no latency improvement.
Distributing model parameters across stages reduces per-GPU memory pressure,
enabling lower power states between micro-batches, an effect entirely invisible
to latency-based reasoning.

Second, a non-monotone energy spike is visible at $(\text{TP}{=}2,
\text{PP}{=}1)$ for all three large models: energy rises relative to its
neighbours even as latency continues to fall. Two-way tensor parallelism keeps
both GPUs in high-utilisation states with frequent all-reduce communication,
whereas configurations involving PP allow pipeline stages to idle between
micro-batches. The result is a configuration that is simultaneously faster and
less energy-efficient than a PP alternative.

\begin{tcolorbox}[colback=blue!5!white, colframe=blue!60!black,
  title=\textbf{Latency and energy optima can diverge.}]
Pipeline parallelism reduces energy by distributing memory pressure across
devices and enabling lower GPU power states an effect
that a latency-based optimiser will systematically overlook.
\end{tcolorbox}

\subsubsection{Cross-Platform Energy Comparison}
\label{subsubsec:cross-platform}

\begin{figure}[htbp]
    \centering
    \includegraphics[width=\columnwidth]{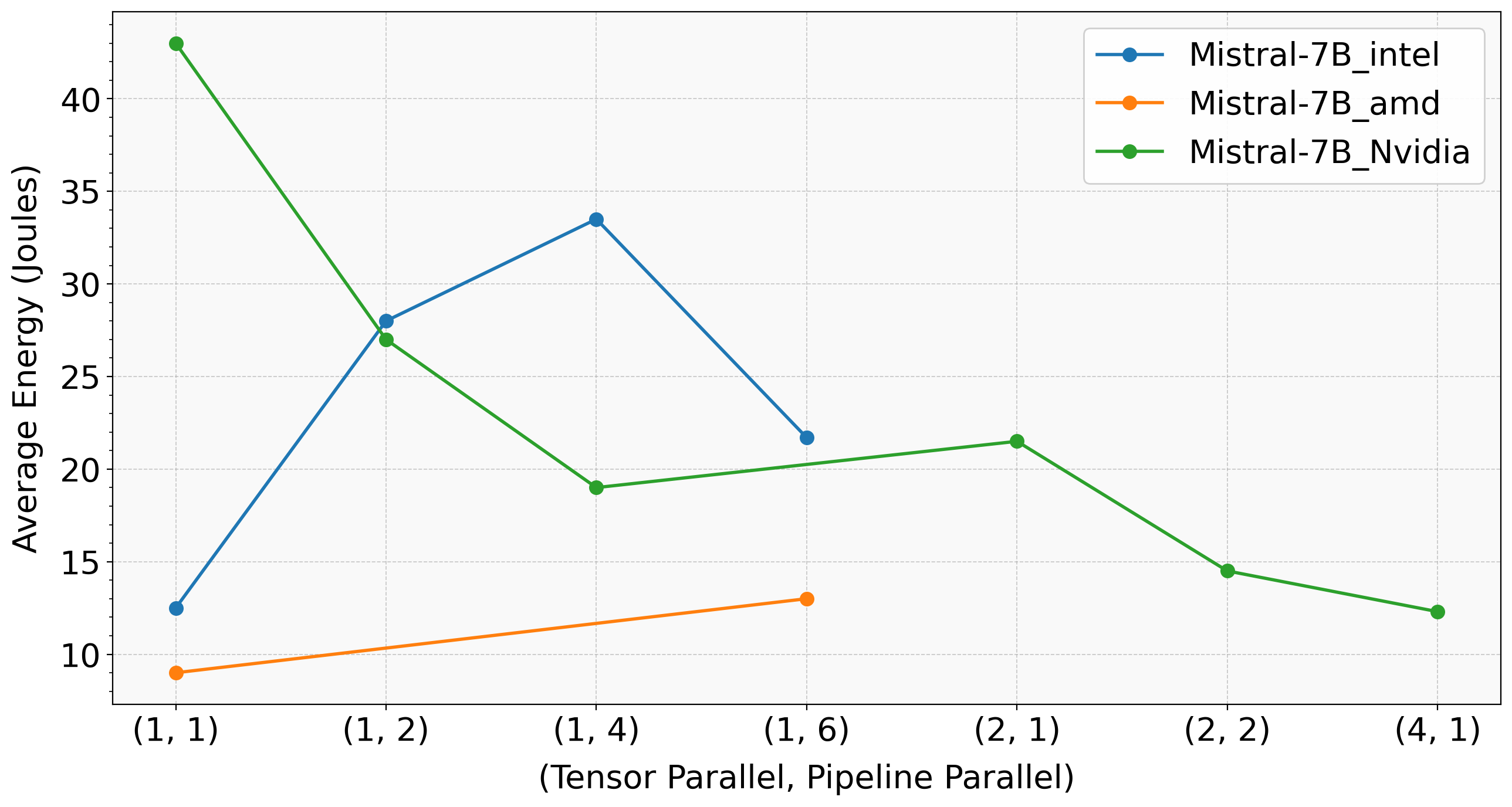}
    \caption{Average energy (Joules) versus TP/PP configuration for Mistral-7B
    across Intel Aurora, AMD MI100, and NVIDIA A100. AMD achieves the lowest
    absolute energy; NVIDIA benefits most from higher TP degrees; Intel
    exhibits a non-monotone response to PP scaling.}
    \label{fig:tp_pp_energy_all_arch}
\end{figure}

Figure~\ref{fig:tp_pp_energy_all_arch} compares per-inference energy for
Mistral-7B across three hardware platforms. The three curves occupy distinct
energy ranges and respond differently to parallelism, reflecting their
underlying memory and interconnect architectures.

AMD MI100 maintains the lowest absolute energy across all configurations. Its
large unified HBM3 capacity accommodates the entire Mistral-7B parameter set
and KV-cache on a single device, eliminating the cross-device communication
overhead that motivates tensor parallelism at this model scale. NVIDIA A100
shows the steepest energy reduction as TP increases, reflecting the efficiency
of NVLink 3.0 for all-reduce operations distributing memory pressure across
multiple GPUs is rewarded with substantial gains at higher TP degrees.

Intel GPU Max 1550 presents the most irregular profile: energy rises through
intermediate PP degrees before recovering at higher values. Xe Link
synchronisation overhead dominates at intermediate pipeline depths but is
eventually outweighed by the memory distribution benefits at higher PP.


\subsubsection{Effect of Batch Size on Energy}
\label{subsubsec:batch_size_energy}

\begin{figure}[htbp]
    \centering
    \includegraphics[width=\columnwidth]{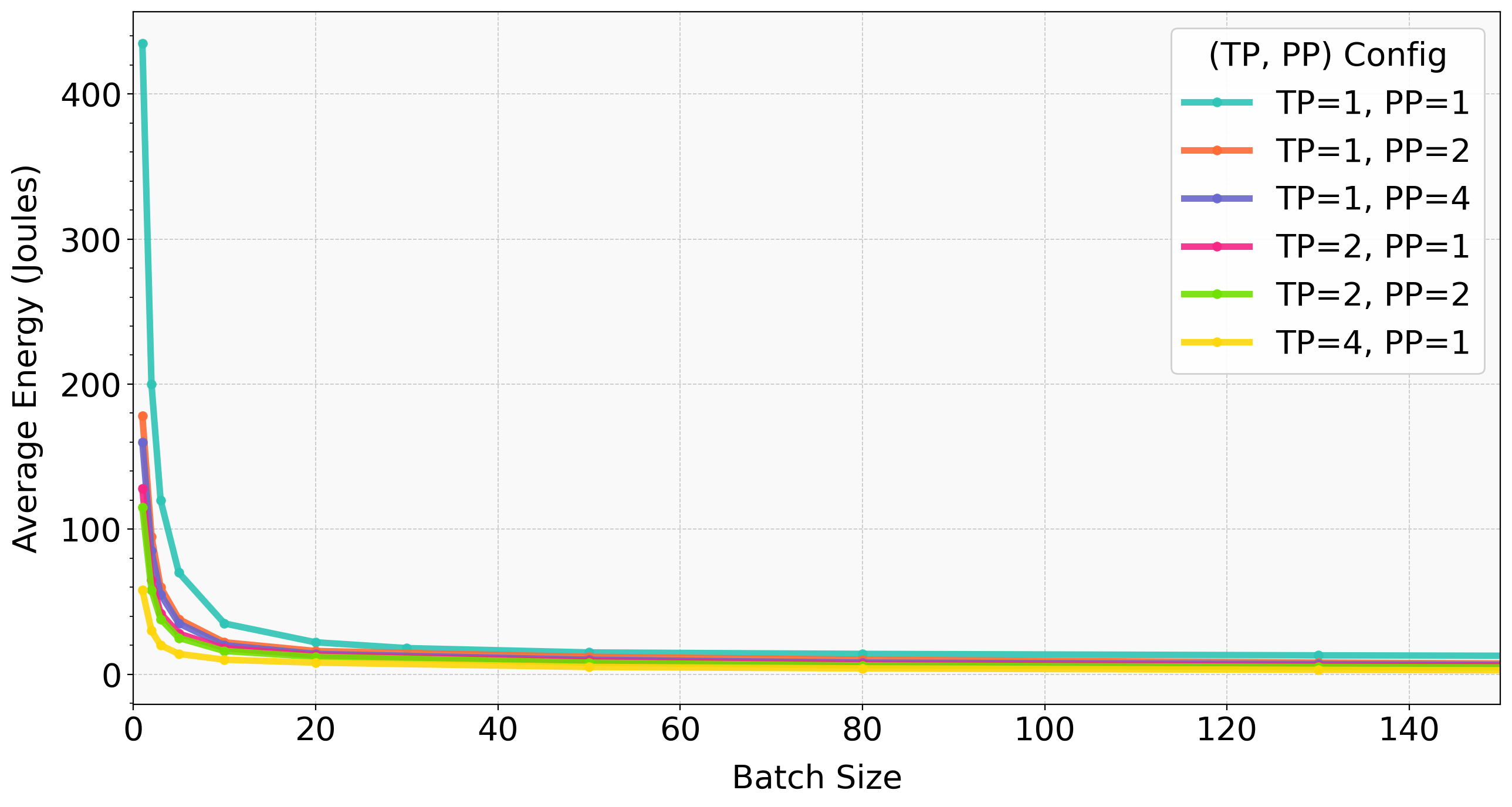}
    \caption{Average per-inference energy (Joules) versus batch size for
    Qwen-7B across TP/PP configurations. All configurations exhibit steep
    energy decay at small batch sizes, converging to $\sim$5-10\,J beyond
    batch size 50. Higher parallelism achieves lower energy at all batch
    sizes.}
    \label{fig:qwen7b_energy_batch_size}
\end{figure}

Figure~\ref{fig:qwen7b_energy_batch_size} shows per-inference energy as a
function of batch size for Qwen-7B under text-only workloads. All curves
follow a sharp decay at small batch sizes before flattening to a near-constant
plateau, with the transition occurring well before batch size 50.

The plateau indicates that fixed overheads model loading, KV-cache
initialisation, kernel launch have been fully amortised, beyond which
further batching yields negligible per-inference savings. The parallelism
hierarchy is preserved across the entire range, but its relative importance
shifts with operating regime: at batch size 1, aggressive parallelism is the
dominant energy lever, while in the saturated regime the advantage narrows,
making simpler configurations increasingly viable.

\begin{tcolorbox}[colback=orange!5!white, colframe=orange!60!black,
  title=\textbf{Batching alone cannot arbitrarily reduce per-inference
  energy.}]
Once fixed costs are amortised, further energy reduction requires optimising
compute-bound operations through architecture changes, quantisation, or more
efficient parallelism not simply increasing batch size.
\end{tcolorbox}

\subsubsection{Effect of Input Modality on Energy}
\label{subsubsec:modality_energy}

\begin{figure}[htbp]
    \centering
    \includegraphics[width=\columnwidth]{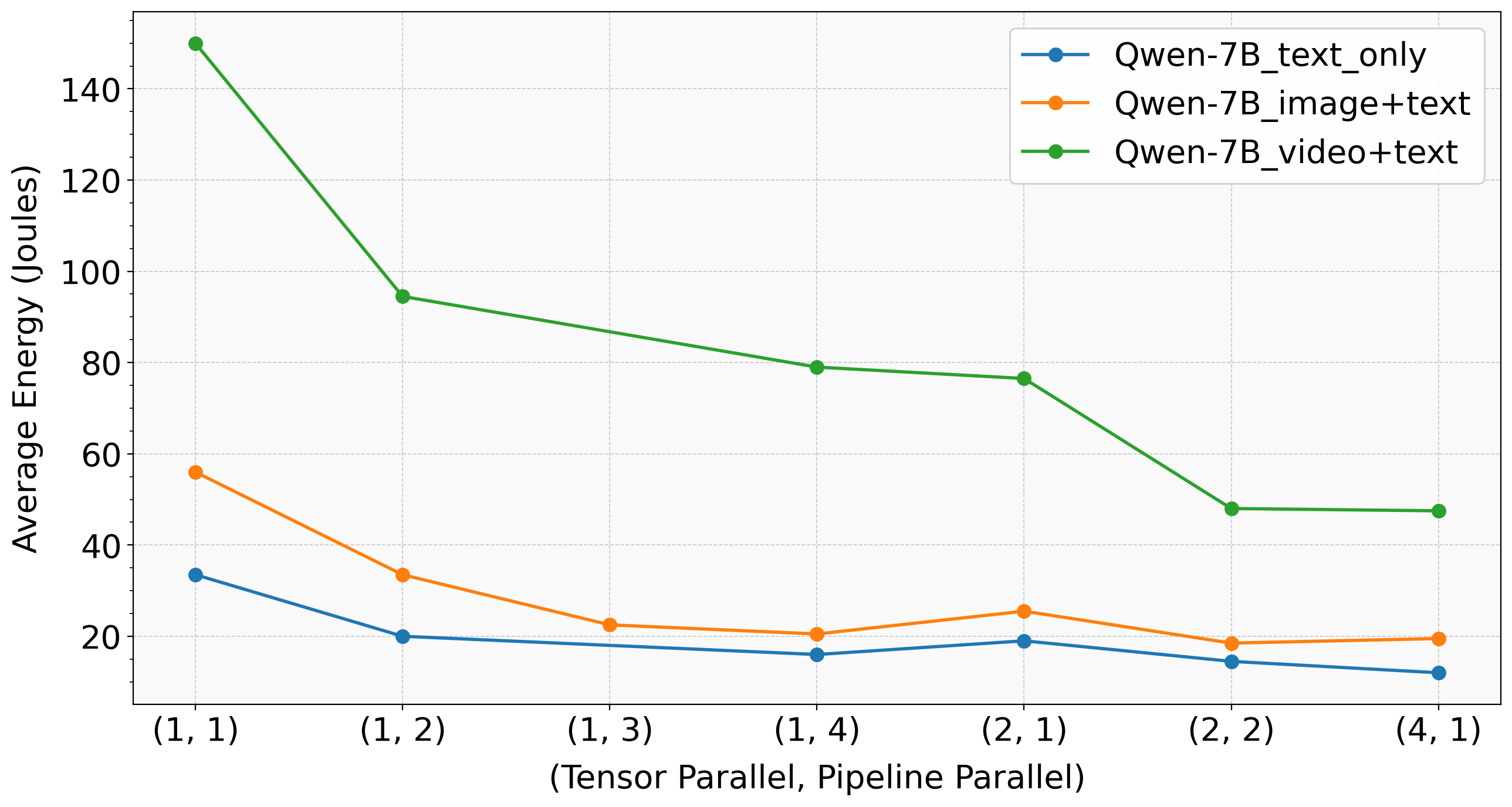}
    \caption{Average energy (Joules) versus TP/PP configuration for Qwen-7B
    under three workload modalities. Video\,+\,text is substantially more
    expensive at baseline; all modalities benefit from parallelism, but
    modality gaps persist even at high TP.}
    \label{fig:qwen7b_energy_tp_pp_modality}
\end{figure}

Figure~\ref{fig:qwen7b_energy_tp_pp_modality} compares energy across text-only,
image\,+\,text, and video\,+\,text inputs for Qwen-7B. The three curves are
vertically separated across the entire TP/PP range, with video\,+\,text
consuming several times more energy than text-only at every configuration
tested.

The gap is most pronounced at low parallelism and compresses but does not
close at high TP, reflecting that the prefill cost of visual tokens dominates
the energy profile and cannot be eliminated through parallelism alone. All
modalities respond strongly to the early pipeline-parallelism steps and all
exhibit the same non-monotone energy increase at $(\text{TP}{=}2,
\text{PP}{=}1)$ observed in the text-only experiments, confirming that
this behaviour is a hardware-level artefact rather than modality-specific.
Unlike text tokens, multimodal prefill cost is per-request and cannot be
amortised across a batch in the same way, making high parallelism a necessity
rather than an option for energy-efficient multimodal serving.

\subsubsection{Effect of Quantization on Energy}
\label{subsubsec:quantization_energy}

\begin{figure}[htbp]
    \centering
    \includegraphics[width=\columnwidth]{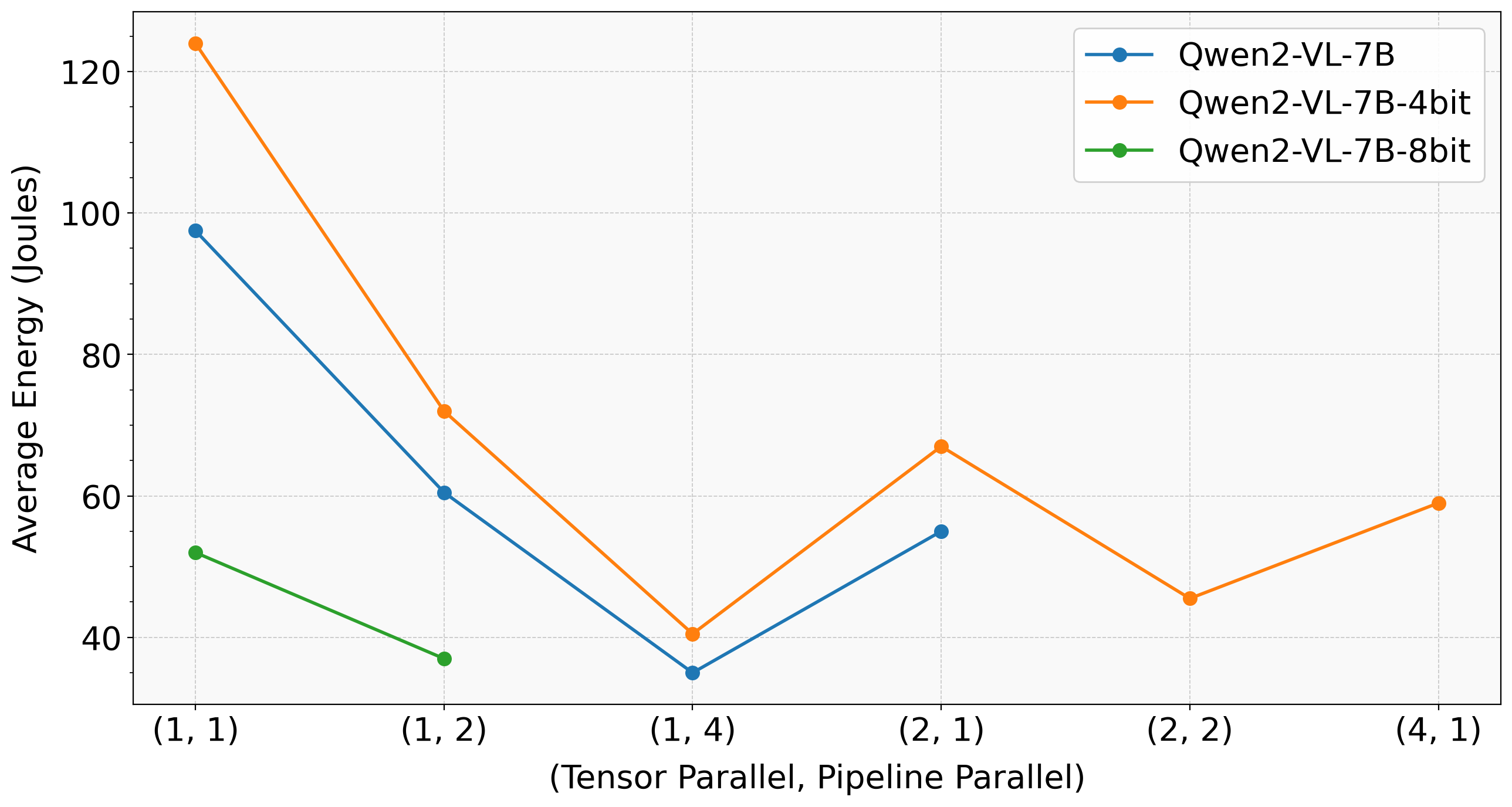}
    \caption{Average energy (Joules) versus TP/PP configuration for
    Qwen2-VL-7B at BF16, 4-bit, and 8-bit quantisation. Counterintuitively,
    4-bit raises energy at most configurations; 8-bit offers the most robust
    advantage.}
    \label{fig:qwen_energy_tp_pp_quant}
\end{figure}

Figure~\ref{fig:qwen_energy_tp_pp_quant} shows how weight quantisation alters
the energy profile of Qwen2-VL-7B. The most striking feature is that 4-bit
quantisation lies \emph{above} the BF16 baseline at low parallelism, even if the intuition would have expected that the reduced precision reduces cost too.

The cause is dequantisation overhead and memory-access irregularity, which
overcome storage savings when GPU resources are limited. This effect is
lost at $(\text{TP}{=}2, \text{PP}{=}1)$, where the 4-bit curve spikes
sharply, indicating a particularly poor interaction between dequantisation
kernels and two-way tensor parallelism. By contrast, 8-bit quantisation
provides consistent energy savings across the measured range and is the more
robust choice. Only at the highest parallelism configurations do the three
curves converge, suggesting that sufficient GPU resources can eventually
compensate for dequantisation overhead.

\begin{tcolorbox}[colback=red!5!white, colframe=red!60!black,
  title=\textbf{Lower bit-width does not imply lower energy.}]
The quantisation-parallelism interaction is non-trivial: 4-bit quantisation
can increase energy at low parallelism due to dequantisation overhead.
Deployments should profile quantised models under their target parallelism
regime rather than assuming bit-width reduction translates directly to energy
reduction.
\end{tcolorbox}

\subsection{Prediction Evaluation}
\label{subsec:predictive-eval}

\subsubsection{Sample Efficiency and Generalization}
\label{subsubsec:sample-efficiency}

Table~\ref{tab:unified-vs-ml-50-500} compares \textsc{EnergyLens} against ML
baselines at $n_{\text{train}}=50$ and $n_{\text{train}}=500$ on Mistral-7B.
Each column reports MAPE, $R^2$, and RMSE; lower MAPE and higher $R^2$
indicate better predictive accuracy.

\textsc{EnergyLens} achieves strong performance already at 50 samples and
improves only marginally at 500, demonstrating that its domain-knowledge
encoding of token, batch, and parallelism interactions allows accurate
parameter estimation from a profiling session that typically completes in
minutes. Data-driven approaches require substantially more data to reach
comparable performance: Random Forest and Gradient Boosting both improve
dramatically between 50 and 500 samples, while Linear Regression remains
poor at both scales due to its inability to capture nonlinear interactions.

\begin{table}[h!]
  \centering
  \caption{%
    MAPE, $R^2$, and RMSE at $n_{\text{train}}=50$ and $n_{\text{train}}=500$
    on Mistral-7B. Bold marks the best value per column.
  }
  \label{tab:unified-vs-ml-50-500}
  \footnotesize
  \setlength{\tabcolsep}{4pt}
  \begin{tabular}{lcccc}
    \toprule
    & \multicolumn{2}{c}{$n=50$}
    & \multicolumn{2}{c}{$n=500$} \\
    \cmidrule(lr){2-3} \cmidrule(lr){4-5}
    & MAPE / $R^2$ & RMSE & MAPE / $R^2$ & RMSE \\
    \midrule
    \textsc{EnergyLens}   & \textbf{13.0\%} / \textbf{0.977} & \textbf{4.41}
                          & \textbf{11.3\%} / \textbf{0.981} & \textbf{4.01} \\
    Linear Reg.           & 127.6\% / 0.409 & 22.54
                          & 100.3\% / 0.476 & 21.23 \\
    Random Forest         & 49.0\% / 0.560  & 19.40
                          & 12.3\% / 0.917  & 8.43  \\
    Grad.\ Boosting       & 38.0\% / 0.483  & 20.38
                          & 13.2\% / 0.941  & 7.11  \\
    \bottomrule
  \end{tabular}
\end{table}

All experiments use random sampling to draw the $n_{\text{train}}$ profiling
measurements. In a controlled ablation over 10 seeds at $n{=}50$, random
sampling consistently outperforms Latin Hypercube Sampling for
\textsc{EnergyLens} (e.g., $13.8\%{\pm}1.3\%$ vs.\ $25.9\%{\pm}1.7\%$ MAPE
on Mistral-7B). The energy surface is steepest at small batch sizes, and
random draws over-represent this high-gradient region naturally, whereas LHS \cite{McKay1979}
spreads samples uniformly and under-samples exactly the region that drives the
batch-size and parallelism exponent fits in Eq.~\eqref{eq:unified-tp-pp}.

\subsubsection{Cross-Hardware and Cross-Model Generalization}
\label{subsubsec:cross-hardware}

Table~\ref{tab:unified-vs-all-50} evaluates \textsc{EnergyLens} alongside all baselines and \textsc{MaverIQ} across multiple hardware--model combinations at $n_{\text{train}}=50$.

\textsc{EnergyLens} is consistently competitive and often achieves the best performance across diverse settings.
It outperforms all baselines, including \textsc{MaverIQ}, on the majority of NVIDIA configurations.
For example, on Nemotron-Nano-9B and Mistral-7B, it achieves the lowest MAPE (15.6\% and 13.0\%, respectively), substantially improving over both ML models and the latency-based proxy.

Across multimodal workloads, \textsc{EnergyLens} continues to perform strongly, achieving the best results on most Qwen2-VL and LLaVA configurations.
There are a few exceptions where data-driven methods slightly outperform it, such as Qwen2-VL video (where Random Forest achieves 28.2\% vs.\ 33.7\%) and LLaVA video (where \textsc{MaverIQ} achieves 49.5\% vs.\ 50.9\%), but these differences remain modest.

On non-NVIDIA hardware, \textsc{EnergyLens} also generalises well without any hardware-specific tuning.
It achieves the best performance on both AMD MI100 (14.8\%) and Intel 1550 (18.1\%), outperforming all baselines.

Overall, these results confirm that the functional form of Eq.~\eqref{eq:unified-tp-pp} generalises effectively across GPU architectures and model families, requiring only the re-fitting of twelve scalar parameters.

\begin{table}[t]
\centering
\footnotesize
\setlength{\tabcolsep}{2pt}
\caption{MAPE of \textsc{EnergyLens}, ML baselines,
and \textsc{MaverIQ}~\cite{maveriq2025} at $n_{\text{train}}=50$ across
hardware and model combinations. Bold marks the best MAPE per row.}
\label{tab:unified-vs-all-50}
\resizebox{\columnwidth}{!}{
\begin{tabular}{lccccc}
\toprule
Modality
& \parbox{0.9cm}{\centering \textsc{Energy-}\\ \textsc{Lens}}
& \parbox{0.9cm}{\centering Lin.\\Reg.}
& \parbox{0.9cm}{\centering Rand.\\Forest}
& \parbox{0.9cm}{\centering Grad.\\Boost.}
& \parbox{0.9cm}{\centering Lat.$\times$P\\(\textsc{MaverIQ})} \\
& MAPE & MAPE & MAPE & MAPE & MAPE \\
\midrule

Nemotron-Nano-9B (NVIDIA) -- Text
& \textbf{15.6\%} & 104.0\% & 53.4\%
& 44.3\% & 36.1\% \\

Mistral-7B (NVIDIA) -- Text
& \textbf{13.0\%} & 127.6\% & 49.0\%
& 38.0\% & 36.2\% \\

Qwen1.5-MoE-2.7B (NVIDIA) -- Text
& \textbf{32.2\%} & 73.9\% & 53.0\%
& 40.0\% & 36.5\% \\

Qwen2-VL-7B (NVIDIA) -- Text
& \textbf{23.3\%} & 111.0\% & 55.8\%
& 48.7\% & 48.5\% \\

Qwen2-VL-7B (NVIDIA) -- Image
& \textbf{28.7\%} & 74.2\% & 50.9\%
& 54.5\% & 56.5\% \\

Qwen2-VL-7B (NVIDIA) -- Video
& 33.7\% & 34.3\% & \textbf{28.2\%}
& 31.8\% & 33.0\% \\

LLaVA-1.5-7B (NVIDIA) -- Image
& \textbf{34.6\%} & 129.5\% & 54.0\%
& 47.9\% & 66.1\% \\

LLaVA-1.5-7B (NVIDIA) -- Video
& 50.9\% & 109.8\% & 60.6\%
& 64.3\% & \textbf{49.5\%} \\

LLaVA-1.5-7B (NVIDIA) -- Text
& \textbf{12.8\%} & 119.6\% & 46.3\%
& 34.5\% & 44.2\% \\

Mistral-7B (AMD MI100) -- Image Chat
& \textbf{14.8\%} & 29.5\% & 18.4\%
& 15.1\% & 23.9\% \\

Mistral-7B (INTEL 1550) -- Video Chat
& \textbf{18.1\%} & 96.8\% & 33.7\%
& 20.2\% & 37.1\% \\

\bottomrule
\end{tabular}
}
\end{table}

\subsubsection{Pairwise Ranking Quality}
\label{subsubsec:pairwise-ranking}

Prediction accuracy (MAPE) measures absolute error, but configuration selection only requires \emph{correct ranking}: if a model consistently identifies which of two configurations uses less energy, it will select near-optimal deployments even when its absolute predictions carry significant error.
We therefore evaluate both \textsc{EnergyLens} and \textsc{MaverIQ} on \textit{pairwise ranking accuracy}: for every pair of $(TP, PP)$ configurations within a scenario, we check whether the model correctly predicts which configuration consumes less energy.
With $k$ configurations per scenario, this yields $\binom{k}{2}$ comparisons 36 pairs for $k{=}9$ providing a far more stable evaluation signal than the winner-take-all overhead metric.
We additionally report Spearman rank correlation~($\rho$) and Top-1 selection accuracy (whether the model's top pick matches the oracle).

Table~\ref{tab:pairwise-ranking} presents results across all nine workloads.
\textsc{EnergyLens} achieves \textbf{95.8\%} pairwise accuracy overall (940 scenarios) compared to \textsc{MaverIQ}'s 91.5\%, with Spearman $\rho = 0.951$ versus 0.896.
Most strikingly, \textsc{EnergyLens} achieves \textbf{88.2\%} Top-1 accuracy correctly identifying the single best configuration in nearly 9 out of 10 scenarios compared to \textsc{MaverIQ}'s 60.9\%.
On text-only workloads, \textsc{EnergyLens} reaches 99.3\% pairwise accuracy on Mistral-7B with a mean regret of just 0.26\%.
Even on the challenging multimodal workloads where absolute MAPE is higher, ranking quality remains strong: 94.6\% on LLava-image-chat and 88.3\% on Qwen2-VL-image-chat, demonstrating that the prefill/decode formula structure captures the correct energy ordering across parallelism configurations even when absolute magnitude predictions carry larger error.

\begin{table}[t]
\centering
\footnotesize
\setlength{\tabcolsep}{3pt}
\caption{%
  Pairwise ranking accuracy, Spearman $\rho$, Top-1 selection accuracy,
  and mean regret for \textsc{EnergyLens} and \textsc{MaverIQ} across
  nine workloads (940 total scenarios). Bold marks the better value per row.
}
\label{tab:pairwise-ranking}
\resizebox{\columnwidth}{!}{
\begin{tabular}{l cc cc cc cc}
\toprule
Modality
& \multicolumn{2}{c}{Pairwise Acc.}
& \multicolumn{2}{c}{Spearman $\rho$}
& \multicolumn{2}{c}{Top-1 Acc.}
& \multicolumn{2}{c}{Mean Regret} \\
\cmidrule(lr){2-3}\cmidrule(lr){4-5}\cmidrule(lr){6-7}\cmidrule(lr){8-9}
& \textsc{EL} & \textsc{Mav}
& \textsc{EL} & \textsc{Mav}
& \textsc{EL} & \textsc{Mav}
& \textsc{EL} & \textsc{Mav} \\
\midrule

Mistral-7B
& \textbf{99.3} & 96.0 & \textbf{0.994} & 0.957 & \textbf{97.6} & 71.7 & \textbf{0.3} & 3.9 \\

Nemotron-V2-9B
& \textbf{96.1} & 90.6 & \textbf{0.953} & 0.877 & \textbf{88.7} & 59.8 & \textbf{2.1} & 7.7 \\

Qwen1.5-MoE
& \textbf{94.8} & 88.1 & \textbf{0.942} & 0.854 & \textbf{84.5} & 46.4 & \textbf{19.0} & 29.2 \\

Qwen2-VL -- Text
& \textbf{95.8} & 92.5 & \textbf{0.953} & 0.915 & \textbf{90.0} & 65.9 & \textbf{3.4} & 6.6 \\

Qwen2-VL -- Image
& 88.3 & \textbf{89.3} & \textbf{0.877} & 0.874 & 45.0 & \textbf{60.0} & 12.5 & \textbf{5.6} \\

Qwen2-VL -- Vid-only
& \textbf{86.7} & 79.2 & \textbf{0.811} & 0.696 & \textbf{68.8} & 31.2 & \textbf{18.2} & 35.8 \\

LLava -- Vid-chat
& 83.3 & \textbf{90.8} & 0.793 & \textbf{0.907} & \textbf{87.5} & 75.0 & \textbf{3.9} & 5.1 \\

LLava -- Img-chat
& \textbf{94.6} & 90.4 & \textbf{0.939} & 0.893 & \textbf{81.2} & 68.8 & \textbf{4.4} & 8.8 \\

\midrule
\textbf{Overall}
& \textbf{95.8} & 91.5 & \textbf{0.951} & 0.896 & \textbf{88.2} & 60.9 & \textbf{7.0} & 12.1 \\

\bottomrule
\end{tabular}
}
\end{table}

\subsubsection{Robustness Without Per-Configuration Power Measurements}
\label{subsubsec:power-degradation}

A practical limitation of latency-based energy proxies, such as \textsc{MaverIQ}, is their reliance on precise, per-configuration power measurements to convert predicted latency into accurate energy estimates.
In production environments, gathering granular power telemetry for every possible hardware configuration is often infeasible.
Consequently, deployments may fall back on a static power estimate such as the mean power consumption across all configurations.

To measure the impact of this limitation, we evaluate the predictive robustness of \textsc{EnergyLens} against \textsc{MaverIQ} (constrained to use mean power) using pairwise ranking accuracy.

\begin{figure}[htbp]
  \centering
  \includegraphics[width=\columnwidth]{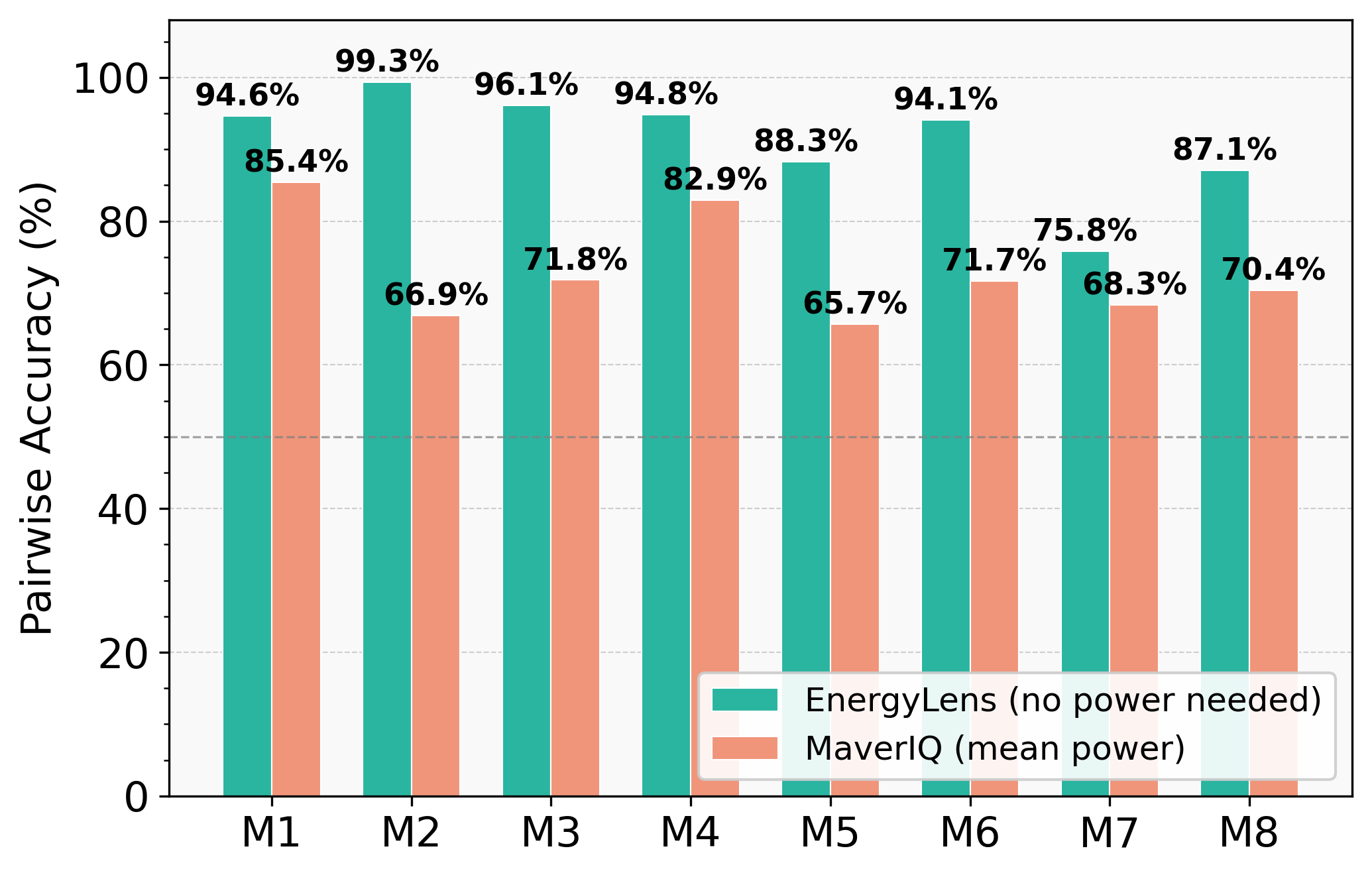}
  \caption{%
    Pairwise accuracy without per-configuration power measurements.
    \textsc{EnergyLens} models energy directly and maintains high accuracy across all text and multimodal tasks.
    In contrast, \textsc{MaverIQ}, when restricted to a mean power proxy, suffers widespread degradation.
    Models: M1~=~LLaVA-1.5-7B-image-chat, M2~=~Mistral-7B, M3~=~NVIDIA-Nemotron-V2-9B,
    M4~=~Qwen1.5-MoE-2.7B, M5~=~Qwen2-VL-7B-image-chat, M6~=~Qwen2-VL-7B-text,
    M7~=~Qwen2-VL-7B-video-chat, M8~=~Qwen2-VL-7B-video-only.
  }
  \label{fig:power_degradation}
\end{figure}

As shown in Figure~\ref{fig:power_degradation}, \textsc{EnergyLens} which models energy natively and requires no external power telemetry exhibits strong predictive robustness across eight distinct workloads.
In contrast, \textsc{MaverIQ} suffers significant degradation in ranking accuracy when deprived of exact power data for each configs.
For instance, on Mistral-7B, \textsc{EnergyLens} achieves a pairwise accuracy of 99.3\%, whereas \textsc{MaverIQ} drops to 66.9\%.
This severe vulnerability is consistent across diverse models: \textsc{EnergyLens} significantly outperforms \textsc{MaverIQ} on NVIDIA-Nemotron-V2-9B (96.1\% vs.\ 71.8\%), Qwen1.5-MoE-2.7B (94.8\% vs.\ 82.9\%), and LLava-1.5-7B-image-chat (94.6\% vs.\ 85.4\%).

The performance gap remains evident even in complex multimodal tasks.
On Qwen2-VL-7B-image-chat, \textsc{MaverIQ}'s accuracy degrades to 65.7\% trending perilously close to the 50\% threshold of random guessing while \textsc{EnergyLens} maintains an 88.3\% accuracy.
These results demonstrate that relying solely on latency as an energy proxy is non optimal approach in real-world deployments; modelling energy directly can provide a vastly superior ability to identify energy-optimal configurations out-of-the-box.

\section{Conclusion and Future works}
\label{sec:conclusion}

We presented \textsc{EnergyLens}, a symbolic-regression framework that derives compact, interpretable closed-form energy formulas for LLM inference from as few as 50 profiling measurements. By decoupling tensor and pipeline parallelism and separating prefill from decode, \textsc{EnergyLens} captures the mechanisms governing GPU energy across diverse model families, modalities, and hardware vendors without structural changes to the formula. Across 940 scenarios spanning dense transformers, MoE, and state-space models on NVIDIA, Intel, and AMD GPUs, it achieves 95.8\% pairwise ranking and 88.2\% Top-1 selection accuracy, versus 91.5\% and 60.9\% for \textsc{MaverIQ}. It matches ensemble ML accuracy with 10$\times$
fewer samples and extrapolates reliably to batch sizes three to ten times larger than those seen during fitting ($\geq$93\%)
pairwise accuracy on five of seven datasets). 

Several limitations point to future work. First, the current template uses additive prefill and decode terms with independent parallelism exponents; allowing multiplicative interactions between TP, PP, and sequence length could improve accuracy on MoE and long-context video workloads where cross-term effects matter. Second, adding domain-specific features such as expert routing frequency or frame-encoding cost could close the accuracy gap on these workloads. Finally, integrating \textsc{EnergyLens} with dynamic schedulers like \textsc{vLLM}'s continuous batching would enable online reconfiguration, moving from static selection to adaptive, energy-aware serving.

We will release \textsc{EnergyLens}  along with the associated datasets, as open-source on GitHub to support reproducibility and foster collaboration within the community.

\bibliographystyle{IEEEtran}
\bibliography{main.bib}

\end{document}